\documentclass[12pt]{article}

\usepackage[utf8]{inputenc}
\usepackage[T1, OT2,OT1]{fontenc}
\usepackage{textcomp}
\usepackage[left={1in},
            right={1in},
            top={1in},
            bottom={1in},
            foot={0.5in}]{geometry}
\usepackage{natbib}
\usepackage{amsmath}
\usepackage{amsthm}
\usepackage{stmaryrd}
\usepackage{xcolor}
\definecolor{black}{rgb}{0,0,0}
\definecolor{splinkcolor}{rgb}{.0,.2,.4}
\usepackage[colorlinks, linkcolor=black, urlcolor=splinkcolor, citecolor=black]{hyperref}
\usepackage{examples-slim}
\usepackage{qtree}
\usepackage{fancyhdr}

\bibpunct[; ]{(}{)}{;}{a}{}{,}  

\newcommand{\secref}[1]{section~\ref{#1}}

\newcommand{\eg}[1]{(\ref{#1})}
\newcommand{\subeg}[2]{(\ref{#1}\ref{#2})}

\newcommand{\dashsubeg}[3]{(\ref{#1}\ref{#2}--\ref{#3})}

\newcommand{\posscitet}[1]{\citeauthor{#1}'s (\citeyear{#1})}

\newcommand{\possciteauthor}[1]{\citeauthor{#1}'s}

\newcommand{\seccitealt}[2]{\citealt{#1}:$\S$#2}
\newcommand{\pgcitep}[2]{(\citealt{#1}:#2)}

\newcommand{\pgcitet}[2]{\citeauthor{#1} (\citeyear{#1}:#2)}

\newcommand{\word}[1]{\emph{#1}}
\newcommand{\tech}[1]{\textsc{#1}}
\newcommand{\emphasis}[1]{\textsc{#1}}

\newcommand{\sem}[1]{\ensuremath{\llbracket#1\rrbracket}}
\newcommand{\semt}[1]{\sem{\text{#1}}}
\newcommand{\semDL}[1]{\semt{#1}^{\text{DL}}}

\newcommand{\activation}{\mathbf{g}}

\newcommand{\fa}[2]{\mathbf{fa}(#1, #2)}

\newcommand{\dlfa}[2]{\activation\left(\left[\semDL{#1};\semDL{#2}\right]W\right)}

\newcommand{\myvec}[2]{\left[\begin{array}{@{} *{#1}{r} @{}} #2 \end{array}\right]}

\qtreecenterfalse

\begin{document}


\title{A case for deep learning in semantics\footnote{My thanks to Sam
    Bowman, Ignacio Cases, Chris Manning, Louise McNally, Joe Pater,
    and Nicholas Tomlin for incisive comments. All views are my
    own.}}%
\author{Christopher Potts}%
\date{Stanford Linguistics}

\maketitle

\thispagestyle{fancy}%
\fancyhead{} %
\lhead{{\footnotesize To appear in \emph{Language} as
  \href{https://journals.linguisticsociety.org/language/index.php/language/pater}{a commentary on Pater 2018}}}
\rhead{{\footnotesize\today}}%
\cfoot{\thepage}
\renewcommand{\headrulewidth}{0pt}


\section{Introduction}\label{sec:intro}

\possciteauthor{Pater:2018} target article builds a persuasive case
for establishing stronger ties between theoretical linguistics and
connectionism -- or, as it's recently been rebranded, \tech{deep
  learning} (DL; \citealt{LeCun:Bengio:Hinton:2015}). In this
commentary, I seek to further support his arguments by extending them
to semantics. It's an exciting time for such discussions: DL is
ascendant, and some of its breakthroughs in natural language
processing (NLP) have come from melding assumptions and techniques
from semantics with those of machine learning. Unfortunately,
linguistic semantics has, to date, been much less influenced by DL
research. I am concerned by this. There is now a large, vibrant,
well-funded community of DL researchers working on compositional
semantics, and semantic theory will suffer if semanticists don't
engage with, and help shape, their research agenda.

When considering a role for DL in semantics, it is worth revisiting
what \citet{Partee95} calls \tech{Lewis's Advice}: `In order to say
what a meaning \emphasis{is}, we may first ask what a meaning
\emphasis{does}, and then find something that does that'
\pgcitep{Lewis70GS}{22}. Lewis and his contemporaries proposed that
\tech{intensional functions} do what meanings do, and higher-order
logic in turn became the field's most important toolkit.  Since then,
semanticists have mostly not revisited this decision, though it has
profoundly influenced how we delimit the field of semantics, which
problems receive attention, and what we regard as an explanation.

A DL-based semantic theory would also try to follow Lewis's Advice,
but it would replace intensional functions with $n$-dimensional arrays
of numbers, and machine learning would likely replace logic as the
most-used toolkit for the field. On this basis, one can build theories
with many of the same properties as those of
\citealt{Lewis70GS}. However, as with intensional functions, this
foundational choice has far-reaching effects on the research
agenda. After sketching what a DL-based semantics would be like, this
commentary tries to identify these effects, both good and bad, with an
eye towards a synthesis of intensional and DL
semantics.\footnote{\label{fn:focus}For reasons of space and alignment
  with \possciteauthor{Pater:2018} article, I focus on DL-based
  computational models. In NLP, \tech{semantic parsing models} learn
  to predict logical forms or denotations using weighted symbolic
  semantic grammars. Compared to DL, this is a less radical way to
  combine semantic theory with machine learning. For discussion, see
  \citealt{Liang:Potts:2015}.}


\section{Semantics with functions or arrays}\label{sec:basics}

Current semantic theories adhere to the \tech{principle of
  compositionality}, which says that the meaning of a non-terminal
syntactic node is determined by the meanings of its child nodes and
the semantic rule used to combine them. If a child node is a lexical
item (terminal node), then its meaning is simply retrieved from the
semantic lexicon.

This formulation says very little about the nature of the meanings or
the combination operations.  As I noted above, though, it is standard
to model the meanings as intensional functions (elements in an
intensional model) and, in modern type-driven semantic theories
\citep{Klein85}, the mode of combination is almost always
\tech{function application}. Thus, let $\sem{\cdot}$ be the
interpretation function, mapping syntactic nodes to intensional
functions, and define $\fa{a}{b} = a(b) \text{ or } b(a)$, whichever
is well-formed. Then the tree in (\ref{t}a) would be interpreted as in
(\ref{t}b).
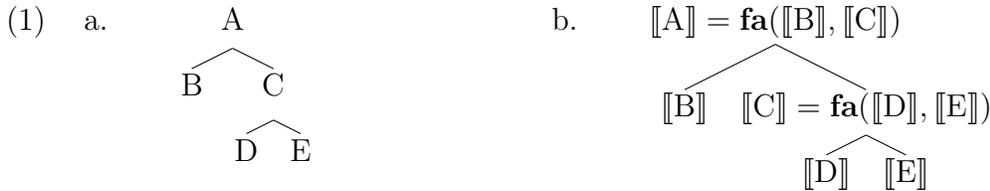
\begin{examples}
\item\label{t}%
  \begin{minipage}[t]{0.4\linewidth}
    \begin{examples}
    \item[a] \Tree[.{A} {B} [.{C} {D} {E} ] ]
    \end{examples}
  \end{minipage}
  \begin{minipage}[t]{0.4\linewidth}
    \begin{examples}
    \item[b] \Tree[.{$\semt{A} = \fa{\semt{B}}{\semt{C}}$}
      {$\semt{B}$} [.{$\semt{C} = \fa{\semt{D}}{\semt{E}}$}
      {$\semt{D}$} {$\semt{E}$} ] ]
    \end{examples}
  \end{minipage}
\end{examples}

A DL-based compositional semantic theory could look very similar to
this. The building blocks for all DL models are $n$-dimensional arrays
of numbers: vectors (one-dimensional), matrices (two-dimensional), and
higher-order tensors. Let $\semDL{$\cdot$}$ be our DL-interpretation
function, mapping syntactic nodes to $n$-dimensional arrays. A
baseline model in this vein would say that lexical items denote
vectors of dimension $m$ and, in place of function application, we
would use the following combination function:
\begin{examples}
\item\label{combo} $f(\semDL{X}, \semDL{Y}) = \dlfa{X}{Y}$
\end{examples}
Here, $W$ is a matrix of dimension $2m \times m$, and $[a;b]$ is the
concatenation of vectors $a$ and $b$, which yields a new vector of
dimension $2m$. The heart of the combination rule is the
multiplication of $[a;b]$ with $W$, which yields a new vector $x$ of
dimension $m$, defined so that
$x_{j} = \sum_{i=1}^{2m} [a;b]_{i}W_{ij}$ Each $x_{j}$ has a
transformation $\activation$ applied to it. For instance, with
$\activation = \tanh$, each of the values is compressed into the range
$(-1, 1)$. This element-wise non-linear transformation gives the
network the power to model very complex functions
\citep{Cybenko:1989,Hornik:1992}. Since the result of all this is
another $m$-dimensional vector, the theory has the desired recursive
property:
\begin{examples}
\item \Tree[.{$\semDL{A} = \dlfa{B}{C}$} {$\semDL{B}$}
  [.{$\semDL{C} = \dlfa{D}{E}$} {$\semDL{D}$} {$\semDL{E}$} ] ]
\end{examples}

To illustrate, suppose we're working in a simplified two-dimensional
semantic space in which the first dimension encodes sentiment valence
and the second encodes intensity.  With the lexical items in
\dashsubeg{lex}{terrible}{not} and the weight matrix $W$ as in
\subeg{lex}{W}, we achieve the effect that \word{not terrible} is less
negative and less intense than \word{terrible} alone, as in
\subeg{lex}{result}.
\begin{examples}
\item\label{lex}%
  \begin{examples}
  \item\label{terrible} $\semDL{terrible} = \myvec{2}{-1.00 & 1.00}$
  \item\label{not} $\semDL{not} = \myvec{2}{-0.10 & 0.10}$
  \item\label{W} $W = \left[
      \begin{array}{r r}
         0.06 &  0.32 \\
        -0.14 & -0.53 \\
         1.24 &  0.00 \\
         0.02 &  1.06 \\
      \end{array}
    \right]$
  \item\label{result}%
    $\semDL{not terrible} = \tanh\left(\left[\semDL{not};
        \semDL{terrible}\right]W\right) = \myvec{2}{-0.85 &  0.75}$
  \end{examples}
\end{examples}

For toy examples like \eg{lex}, we can invent values that achieve the
desired outcome. However, such hand-construction is infeasible for
large, complex semantic systems. In practice, the lexical
representations in \dashsubeg{lex}{terrible}{not} and the weight
matrix $W$ in \subeg{lex}{W} are model parameters that are learned
from data. For the most part, this is a \tech{supervised learning}
process \citep{Hastie:Tibshirani:Friedman:2009}: individual examples
are labeled, and the optimization process finds parameter values that
accurately predict those labels. To scale up \eg{lex}, for example, we
might collect real-valued polarity and intensity judgments for a wide
range of adverb--adjective pairs, learn lexical entries and weights to
predict those labels, and then evaluate the model by how accurately it
can predict the labels for combinations it didn't see during
optimization. In DL, these labels can take many forms and come from
many different kinds of information; the ambition of the field is that
they will ultimately be dense, high-dimensional representations of
complex environments (which could be physical spaces, simulated
worlds, large multi-media databases, and so forth). The recent surge in
interest in DL derives in large part from advances in machine learning
and computing power that together make it possible to optimize even
very elaborate models effectively.

\citet{Socher-etal:2011:EMNLP} pioneered the use of tree-structured
neural models for semantic tasks (see also
\citealt{Socher-etal:2012,Socher-etal:2013}), building on early work
on recursive connectionist architectures
\citep{Pollack:1990,Smolensky:1990,Plate:1994, Goller:Kuchler:1996}
and on compositional distributional semantics
(\citealt{Mitchell:Lapata:2010, Baroni:Zamparelli:2010}; for an
overview, see \citealt{Baroni-etal:2014}). These proposals are
explicitly guided by the principle of compositionality and the usual
assumptions and practices of formal semantics. Recurrent neural
networks (RNNs; \citealt{Elman:1990}) are closely related variants in
which the tree structure is a sequence (a strictly right-branching
tree; see \seccitealt{Pater:2018}{4.2}). RNNs make fewer assumptions
about what the incoming data are like than tree-structured
architectures, but they still have the capacity to model aspects of
semantic composition because each non-terminal node is represented in
part by a lexical item and in part by a representation of the
preceding sequence. RNNs and their variants are currently the most
widely used DL architectures for language tasks.\footnote{I see two
  angles on the preference for RNNs over more richly structured
  models. It could derive from what \citeauthor{Pater:2018} calls the
  `emergentist tradition' of the field, which favors powerful models
  that make few assumptions about the data prior to learning. However,
  I suspect DL researchers would eagerly adopt tree-structured models
  if they showed consistent benefits, but so far they have not. It
  would be hasty to conclude that this tells us language isn't
  tree-structured, though. It's safer to conclude that the tree
  structures we're assuming are simply incorrect enough that they get
  in the way. Data-driven techniques like those of DL could help us
  discover the right trees.}

Those are the basics. I henceforth use the labels \tech{intensional
  semantics} and \tech{DL semantics} to refer to the two general
theoretical frameworks exemplified in the above sketches.  These
labels are unfairly reductionist, but I like how the first emphasizes
how meanings are reconstructed and the second emphasizes the role of
machine learning. The frameworks differ along other dimensions (e.g.,
logical vs.~computational, symbolic vs.~numerical), but I think these
differences are largely incidental by comparison.


\section{Learning and usage}\label{sec:learning}

Learning is the crux of \posscitet{Pater:2018} arguments in favor of
combining DL and generative approaches to language. Those issues might
loom even larger in the context of semantics, which has long wrestled
with the tension that \pgcitet{Partee:1980}{1} identifies in the
following passage (\citealt{Montague74} is a foundational collection,
the basis for what came to be known as `Montague Grammar'):
\begin{quote}
  The view that semantics is a branch of psychology is a part of the
  Chomskyan view that linguistics as a whole is a branch of
  psychology.
  [\ldots]
  The contrasting view is ascribed to (and endorsed) by Thomason in
  his introduction to \citealt{Montague74}: `Many linguists may not
  realize at first glance how fundamentally Montague's approach
  differs from current linguistic conceptions. According to Montague
  the syntax, semantics, and pragmatics of natural languages are
  branches of mathematics, not of psychology.'
\end{quote}
The fact that DL models learn lexical entries and combination rules
from data has profound effects on the resulting theories of meaning.
With only a very few exceptions (like learning from dictionaries), the
data set used to optimize the system will be, in one way or another, a
record of \emphasis{utterances} rather than idealized linguistic
objects. As a result, it will reflect many aspects of language use:
biases in word frequency, preferences for certain readings, pragmatic
refinements of lexical items, and so forth. All of these factors will
likely make their way into the final theory, in the sense that the
learned model will reproduce these usage patterns. In other words, it
will not only \emphasis{represent} meanings, it will also make
predictions about the interpretive choices listeners are likely to
make when the language is used for some purpose (whatever purpose
guided the data collection and labeling).

It seems clear that these consequences of learning compromise another
methodological edict of \pgcitet{Lewis70GS}{19}:
\begin{quote}
  I distinguish two topics: first, the description of possible
  languages or grammars as abstract semantic systems whereby symbols
  are associated with aspects of the world; and second, the
  description of the psychological and sociological facts whereby a
  particular one of these abstract semantic systems is the one used by
  a person or population. Only confusion comes of mixing these two
  topics.
\end{quote}
Now, this edict \emphasis{could} lead researchers to seek out
innovative ways in which to abstract out an idealized semantics from
the usage patterns encoded in a DL model's learned parameters. This
would provide rich new perspectives on lexicography, on how children
acquire semantic content from experience
\citep{Frank:Goodman:Tenenbaum:2009}, and on why some aspects of
semantic content are themselves highly variable and uncertain
\citep{Clark97,Wilson:Carston:2007,Potts:Levy:2015}.

However, the dominant reaction from the NLP community has been to
accept that their systems blur meaning and use together.  If one wants
one's system to do interesting things in the real world, it is
generally desirable to have it, for example, resolve ambiguities
rather than simply representing them. A DL semantics will likely only
further encourage the blurring of these boundaries, since DL makes it
easy to combine diverse representations -- of language, of the
physical environment, of others' mental states -- and learn from how
these representations interact. In this setting, it can be hard to
discern a motivation for the restrictive view that Lewis advocates
above. A DL semantics will naturally encourage holistic exploration of
the full significance of utterances, with narrow semantic
characterizations employed only where they prove useful for meeting
these broader goals. Certain insights and goals might be subverted in
the process.

To take one simple example, \citet{Socher-etal:2013} motivate a
tree-structured recursive neural tensor network, which is essentially
an elaboration of the model in \eg{combo} with a greater capacity to
capture the relationships between the two child vectors. Their
experiments are conducted on a corpus of sentences from movie reviews
labeled at the phrase- and sentence-level for their evaluative
sentiment. One of the case studies they report concerns coordination
with \word{but}. They make the case that the learned model reflects
the generalization, due to \citet{RLakoff:1971}, that \word{A but B}
concedes that A and argues that B, because the model consistently
predicts that the sentiment of \word{A but B} is largely determined by
the sentiment of \word{B}. This strikes me as an innovative way to use
a learned DL model to support a nuanced generalization about
meaning. However, there is no clear sense in which the model also
captures what semanticists might regard as the essence of this lexical
item: it is a coordinator that conveys a secondary meaning (perhaps a
\tech{conventional implicature}) that has its own compositional
properties.

All of this makes salient the role that intensional functions and
their attendant logical apparatus play in encouraging the sort of
division that Lewis advocates for above. Logics are the paradigm cases
of closed, self-contained formal systems defined independently of
particular users or instances of use. When we embed our semantic
theories in these systems, the theories inherit these properties. This
has undoubtedly had an effect on the sort of phenomena that linguistic
semanticists choose to study. It separates semantics from all aspects
of learning and cognitive representation
\citep{Partee:1980,Partee81,Jackendoff96}, and it naturally
discourages work on items that are explicitly tied to interactional
language use -- disfluencies, swears, honorifics, interjections, and
other items that \citet{Kaplan99} characterizes in terms of their
\tech{use conditions}. Where such items are studied, it tends to be
from the perspective of how they are represented model-theoretically
and how they interact with the rest of the compositional system,
rather than from the perspective of what they actually mean (in the
pre-theoretic sense) when speakers actually use them. Just as it might
seem idiosyncratic that \citet{Socher-etal:2013} model only the
argumentative structure of \word{but}, these accounts can also seem
idiosyncratic in how they attend only to what intensional semantics
can easily accommodate.

There is a related, subtler difference worth bringing out. Since any
learning-based theory will depend on data sets of utterances, the
question arises whether these utterances have a unique intended
semantic interpretation. Where there are ambiguities, are they always
resolved, or might both speakers and listeners entertain multiple
possibilities \citep{Clark97}? This issue simply doesn't arise for
classical semantic theories, which confine themselves to the question
of which representations are possible. A learned theory is likely to
rank possible construals, inviting the question of whether the full
ranking has communicative or cognitie significance. In other words,
the tools of DL are leading us to lose track of the distinction
between sentence and utterance, just as intensional theories tend to
force it upon us.


\section{Compositionality and generalization}\label{sec:compositionality}

Both of the theories sketched in \secref{sec:basics} are compositional
in the technical sense originally defined by \citet{Montague70UG}.
However, the weight matrix $W$ in the DL semantics might be seen as
compromising the spirit of compositionality because of its global
character. Not only is $W$ used in every phrasal combination, but its
values are learned from the entire data set. As a result, $W$ can
import language-wide information into the local computation of phrasal
meaning. Relatedly, it can have the effect of spreading information
out across different components of the system. The example in \eg{lex}
begins to suggest how this can happen: while $\semDL{not}$ contributes
to the final values, one can make a case that it is $W$, rather than
$\semDL{not}$, that encodes the core effect of negation.

It would be easy to dismiss these considerations on the grounds that
the compositionality principle is not intended as a statement about
how any agent, human or artificial, would learn the semantics of a
system. We expect such learning processes to be more holistic. The
compositionality principle is meant instead to constrain the final
state of the semantic grammar, and here our DL version passes muster.

Still, these considerations should lead us to reflect on the broader
rationale for the compositionality principle. The usual story is that
compositionality is crucial to our ability to produce and understand
creative new combinations of linguistic units, because it offers
guarantees about the systematicity and predictability of new
units. However, these observations alone do not imply
compositionality. The interpretation of a given phrase could be
systematic, predictable, and also determined in part by global
properties of the utterance, the speaker, the discourse situation, and
so forth. And, indeed, it seems to me that our everyday experiences
with language are in keeping with this. Listeners greedily use all
sorts of information when making sense of others' utterances, and
speakers assume they will do this.

In machine learning, by contrast, the goal is not compositionality per
se, but rather \emphasis{generalization}: a system succeeds to the
degree that it makes good predictions for entirely new data -- to the
extent that it displays a human-like ability to creatively produce and
consume utterances that are novel in the sense that they are not
included in the training data. Compositionality is a highly
restrictive strategy for achieving this. If one is designing a system
with generalization in mind, one is unlikely to restrict access to
potentially useful information a priori. Doing so could weaken the
system and, in any case, modern machine learning models learn which
pieces of information to pay attention to as part of optimization, so
it rarely makes sense to deny them available information during
learning.

Finally, there is another dimension to the contrast between
compositionality and generalization. As \pgcitet{Janssen97}{461}
observes, `Compositionality is not a formal restriction on what can
be achieved, but a methodology on how to proceed' (see also
\seccitealt{Partee84}{7.5}). As a result, though it has had a
meaningful impact on the accounts semanticists develop, it can only do
so much, since we can make pretty much any analysis compositional if
we feel pressed to. In contrast, generalization is something we can
measure quite precisely using quantitative metrics and specially
created data sets used only for assessment. In turn, NLP (like the
rest of artificial intelligence) is driven almost entirely by
quantitative performance. The effects are both good and bad. On the
one hand, even really unusual theories can get a hearing if they post
good numbers, whereas unusual linguistic theories can have trouble
getting a fair hearing. On the other hand, scientific goals can seem
less important than incremental gains on accepted test sets. A truly
interdisciplinary DL semantics would, I think, have a chance of
finding a balance.


\section{Lexical semantics}\label{sec:lexical}

The area of semantics in which DL would have the largest impact is
arguably lexical semantics, and I think it's here that we see most
clearly what the trade-offs would be as compared to intensional
approaches.

In the DL theory sketched in \secref{sec:basics}, lexical items (and,
indeed, all meanings) are vectors. The lexicon is in turn a matrix in
which the rows are lexical items and the columns capture specific
aspects of meaning. One could, in these terms, recapitulate all of
intensional semantics: the columns could represent possible worlds,
with binary vectors encoding truth in those worlds. This would give us
representations for proposition-denoting lexical items.  Higher-order
tensors could capture the dependencies of other arguments, thereby
recreating the typed semantic hierarchy of intransitive verbs,
transitive verbs, prepositions, and so forth. However, this is
probably not the best use of DL's building blocks. Instead, these
representations are more fruitfully used to capture specific
dimensions of semantic meaning, in exactly the same way that, for
example, phonological segments are modeled as binary vectors in which
each dimension corresponds to a feature and $1$ means the feature is
present/true and $0$ means it is absent/false. Such feature
representations are common in many areas of linguistics, and were a
mainstay of generative semantics, so these ideas are perhaps not so
unfamiliar.

What \emphasis{is} unfamiliar about DL modeling of the lexicon is that one
rarely has a solid intuitive grasp on what the dimensions in the
lexical entries mean. They are typically learned from data via a
complicated model, and they are too big and complex to understand
analytically. The meaning they encode is largely latent in the
relational structure of the full lexical space
\citep{TurneyPantel10,Lenci:2018}; a further modeling process might
reveal that the representations can be used to make accurate
predictions about, say, lexical entailment, synonymy, and antonymy,
but this is rarely evident from just looking at them.  This lack of
interpretability is not an inevitable outcome of DL modeling --
\citeauthor{Pater:2018} reviews many cases in which linguists' usual
representations are used as the input to DL models to good effect --
but it is a likely consequence of truly embracing what DL has to offer
the lexical semanticist.

Is this a viable alternative to current theories of the lexicon in
semantics? To address this question, I feel we have to confront the
fact that the lexicon has largely been neglected by modern semantic
theories that find their origins in the work of \citet{Lewis70GS} and
\citet{Montague74}. The foreword to \citet{Carlson77} is a send-up of
this limitation; then a graduate student, \citeauthor{Carlson77} asked
his professors the meaning of \word{life} and was told that it was, in
essence, an atomic and unanalyzed formal symbol \textbf{life}, and
`the class then turned to the much stickier problem of pronouns'. In
introducing the papers in \citet{Montague74},
\pgcitet{Thomason74INTRO}{48--49} is more direct:

\begin{quote}
  The problems of a semantic theory should be distinguished from those
  of lexicography [\ldots] A central goal of (semantics) is to explain
  how different kinds of meanings attach to different syntactic
  categories; another is to explain how the meanings of phrases depend
  on those of their components. [\ldots] But we should not expect a
  semantic theory to furnish an account of how any two expressions
  belonging to the same syntactic category differ in meaning. `Walk'
  and `run,' for instance, and `unicorn' and `zebra' certainly
  do differ in meaning, and we require a dictionary of English to tell
  us how. But the making of a dictionary demands considerable
  knowledge of the world.
\end{quote}

I would argue that this doesn't reflect a true division between
semantics and lexicography, or an intrinsic limitation of semantics.
Rather, it is another case of the tools shaping the theory.  The tools
developed in \citealt{Montague74} are ideally suited to modeling the
functional elements of the vocabulary, but they are cumbersome when
applied to open-class lexical items. Of course, the meanings can in
principle be \emphasis{represented} by these logical theories, but
when it comes to actually building a lexicon, they are mostly no help
at all, and the successful large-scale lexical projects do not use
them, opting instead for capturing lexical knowledge in large graphs
\citep{WordNet98,FrameNet2}.

So the promise of DL semantics is that it will allow us to learn rich
representations of the entire lexicon, including open-domain items
like \word{walk} and \word{zebra}. We can learn such representations
from co-occurrence patterns, from visual images, from grounded
interactional scenarios, and so forth. The important caveat here is
that the representations are unlikely to admit of analytic
understanding.  We might be able to probe them in various ways and, in
doing so, assess whether they have captured a specific set of
properties or constraints, but this is unlikely to be evident from
high-level inspection or straightforward, exact calculations. It
should be said that, given the complexity of natural language
lexicons, these analytic limitations might be unavoidable if the goal
is a truly comprehensive treatment.

It is worth noting also that, echoing themes of \secref{sec:learning},
DL lexical theories discourage firm boundaries between semantics and
pragmatics. The information used to create the lexical representations
is mostly a record of language use, and the results tend to reflect
this. DL theories generally cannot distinguish between literal and
non-literal language use, or between denotation and connotation, or
between semantic content and pragmatic enrichment. As a result, since
these phenomena are pervasive, they take centerstage in a way that
they rarely do in intensional theories.

On the other hand, DL theories have been much less successful to date
in modeling the functional vocabulary that semanticists have mostly
specialized in. This is not a representational challenge, as
\citet{ClarkCoeckeSadrzadeh2011} show with their compositional
distributed models of meaning, but rather one arising from the demands
of machine learning. For instance, the DL theory of
\secref{sec:basics} is essentially \tech{monotyped}: every meaning,
whether lexical or phrasal, is a vector. Semanticists will immediately
see that this is untenable; quantificational determiner meanings are
more complex than common noun meanings, so any theory that puts them
in the same meaning space is unlikely to do justice to
determiners. This can of course have deep consequences for the success
of a model, in ways that only careful linguistic argumentation can
reliably bring out. For instance, \citet{Bowman:2016} shows that even
highly effective DL models trained to do the task of \tech{natural
  language inference} (commonsense reasoning about entailment and
contradiction) can fail to fully capture the monotonicity properties
of quantifiers, which has immediately evident consequences for them as
systems that are supposed to reason in language.


\section{Looking ahead}\label{sec:conclusion}

In a commentary called `Computational linguistics and deep learning',
\pgcitet{Manning:2015}{701} quotes the machine learning researcher
Neil Lawrence as saying, `NLP is kind of like a rabbit in the
headlights of the Deep Learning machine, waiting to be flattened'. At
the time, there was a great deal of optimism about DL throughout the
field of artificial intelligence, and prominent DL researchers often
described language as the next big area for applying DL tools. In this
context, Lawrence's warning invokes an unflattering but common trope:
scientists failing to adapt to revolutionary ideas because of an
irrational commitment to old questions and techniques. Linguists are
especially familiar with this narrative; the story of the Chomskyan
revolution, as told in work like \citet{Searle:1972} and
\citet{Harris93}, is one of intransigent American structural linguists
who were flattened by the Cognitive Science machine.

Lawrence's remark is directed at NLP, not theoretical linguistics, but
we should still ask whether we are at risk of being flattened by the
DL machine as well. As recently as fifteen years ago, NLP semantics
was mainly lexical semantics and heuristic shallow semantic
representation. Semanticists could rest easy that they were the only
ones paying attention to compositional aspects of meaning. This began
to change in about 2005, with an outpouring of research on learned
semantic parsing systems (see footnote~\ref{fn:focus}). The rise of DL
over the past decade has greatly accelerated this change, so that
serious semantic interpretation is now the norm in NLP, in the sense
that almost all researchers use recursive models that represent
meaning compositionally. In my experience, the NLP community is also
admirably self-critical, apt to dwell on where its models are failing
as a way of moving forward. The shortcomings of DL for linguistic
analysis are constantly discussed and debated, and they might
eventually lead to DL being supplanted in some sense. Deep semantic
analysis isn't going anywhere, though -- it's too obviously essential
to achieving systems that can produce and interpret language robustly.

What role will formal semanticists play in this new era of deep,
learned semantics? It is not a foregone conclusion that the results,
values, and methods of formal semantics will survive. Semanticists
will have to insert themselves into the DL discourse to make that
happen. However, I'd like to avoid this negative framing as much as
possible. As our best selves, we can set aside our past theoretical
commitments and current methodological attachments and just pose the
question of whether DL provides tools, ideas, and insights that could
benefit our field. \possciteauthor{Pater:2018} target article gives us
many reasons to say yes, and I think the above discussion does as
well. DL appears outwardly to be driven entirely by engineering
concerns, but its connectionist roots are still strong and, as
\citeauthor{Pater:2018} documents, connectionism was founded on many
of the same core principles as modern linguistics. Both fields can
achieve more if they work together.


\setlength{\bibsep}{4pt}
\bibliography{pater-commentary-by-potts-bib}

\newcommand{\SortNoop}[1]{}
\begin{thebibliography}{44}
\providecommand{\natexlab}[1]{#1}
\providecommand{\url}{\relax}
\providecommand{\urlprefix}{Online: }

\bibitem[{Baroni et~al.(2014)Baroni and colleagues}]{Baroni-etal:2014}
\textsc{Baroni, Marco}; \textsc{Raffaella Bernardi}; and \textsc{Roberto
  Zamparelli}. 2014.
\newblock Frege in space: A program for compositional distributional semantics.
\newblock \emph{Linguistic Issues in Language Technology} 9.241--346.

\bibitem[{Baroni \& Zamparelli(2010)Baroni and
  Zamparelli}]{Baroni:Zamparelli:2010}
\textsc{Baroni, Marco}, and \textsc{Roberto Zamparelli}. 2010.
\newblock Nouns are vectors, adjectives are matrices: Representing
  adjective--noun constructions in semantic space.
\newblock \emph{Proceedings of the 2010 {C}onference on {E}mpirical {M}ethods
  in {N}atural {L}anguage {P}rocessing}, 1183--1193. Cambridge, MA: Association
  for Computational Linguistics.
\urlprefix\url{http://aclweb.org/anthology/D10-1115}.

\bibitem[{Bowman(2016)}]{Bowman:2016}
\textsc{Bowman, Samuel~R.} 2016.
\newblock \emph{Modeling natural language semantics in learned
  representations}.
\newblock Stanford, CA: Stanford University dissertation.

\bibitem[{Carlson(1977)}]{Carlson77}
\textsc{Carlson, Gregory}. 1977.
\newblock \emph{Reference to kinds in {E}nglish}.
\newblock Amherst, MA: UMass Amherst dissertation.

\bibitem[{Clark(1997)}]{Clark97}
\textsc{Clark, Herbert~H.} 1997.
\newblock Dogmas of understanding.
\newblock \emph{Discourse Processes} 23.567--59.

\bibitem[{Clark et~al.(2011)Clark and colleagues}]{ClarkCoeckeSadrzadeh2011}
\textsc{Clark, Stephen}; \textsc{Bob Coecke}; and \textsc{Mehrnoosh Sadrzadeh}.
  2011.
\newblock Mathematical foundations for a compositional distributed model of
  meaning.
\newblock \emph{Linguistic Analysis} 36.345--384.

\bibitem[{Cybenko(1989)}]{Cybenko:1989}
\textsc{Cybenko, George}. 1989.
\newblock Approximation by superpositions of a sigmoidal function.
\newblock \emph{Mathematics of Control, Signals and Systems} 2.303--314.

\bibitem[{Elman(1990)}]{Elman:1990}
\textsc{Elman, Jeffrey~L.} 1990.
\newblock Finding structure in time.
\newblock \emph{Cognitive Science} 14.179--211.

\bibitem[{Fellbaum(1998)}]{WordNet98}
\textsc{Fellbaum, Christiane} (ed.) 1998.
\newblock \emph{Word{N}et: An electronic database}.
\newblock Cambridge, MA: MIT Press.

\bibitem[{Frank et~al.(2009)Frank and
  colleagues}]{Frank:Goodman:Tenenbaum:2009}
\textsc{Frank, Michael~C.}; \textsc{Noah~D. Goodman}; and \textsc{Joshua~B.
  Tenenbaum}. 2009.
\newblock Using speakers' referential intentions to model early
  cross-situational word learning.
\newblock \emph{Psychological Science} 20.578--585.

\bibitem[{Goller \& K{\"u}chler(1996)Goller and
  K{\"u}chler}]{Goller:Kuchler:1996}
\textsc{Goller, Christoph}, and \textsc{Andreas K{\"u}chler}. 1996.
\newblock Learning task-dependent distributed representations by
  backpropagation through structure.
\newblock \emph{{IEEE} {I}nternational {C}onference on {N}eural {N}etworks},
  347--352. IEEE.

\bibitem[{Harris(1993)}]{Harris93}
\textsc{Harris, Randy~Allen}. 1993.
\newblock \emph{The linguistic wars}.
\newblock Oxford: Oxford University Press.

\bibitem[{Hastie et~al.(2009)Hastie and
  colleagues}]{Hastie:Tibshirani:Friedman:2009}
\textsc{Hastie, Trevor}; \textsc{Robert Tibshirani}; and \textsc{Jerome
  Friedman}. 2009.
\newblock \emph{The elements of statistical learning}.
\newblock 2 edn. Berlin: Springer.

\bibitem[{Hornik(1992)}]{Hornik:1992}
\textsc{Hornik, Kurt}. 1992.
\newblock Approximation capabilities of multilayer feedforward networks.
\newblock \emph{Neural Networks} 4.251--257.

\bibitem[{Jackendoff(1996)}]{Jackendoff96}
\textsc{Jackendoff, Ray~S.} 1996.
\newblock Semantics and cognition.
\newblock \emph{The handbook of contemporary semantic theory}, ed.~by Shalom
  Lappin, 539--559. Oxford: Blackwell Publishers.

\bibitem[{Janssen(1997)}]{Janssen97}
\textsc{Janssen, Theo M.~V.} 1997.
\newblock Compositionality.
\newblock \emph{Handbook of logic and language}, ed.~by Johan
  {\SortNoop{Benthem}}van~Benthem and Alice ter Meulen, 417--473. Cambridge, MA
  and Amsterdam: MIT Press and North-Holland.

\bibitem[{Kaplan(1999)}]{Kaplan99}
\textsc{Kaplan, David}. 1999.
\newblock What is meaning? {E}xplorations in the theory of \emph{Meaning as
  Use}. {B}rief version --- draft 1.
\newblock {Ms.}, UCLA.

\bibitem[{Klein \& Sag(1985)Klein and Sag}]{Klein85}
\textsc{Klein, Ewan}, and \textsc{Ivan~A. Sag}. 1985.
\newblock Type-driven translation.
\newblock \emph{Linguistics and Philosophy} 8.163--201.

\bibitem[{Lakoff(1971)}]{RLakoff:1971}
\textsc{Lakoff, Robin}. 1971.
\newblock If's, and's, and but's about conjunction.
\newblock \emph{Studies in linguistic semantics}, ed.~by Charles~J. Fillmore
  and D.~Terence Langendoen, 114--149. New York: Holt, Rinehart, and Winston.

\bibitem[{LeCun et~al.(2015)LeCun and colleagues}]{LeCun:Bengio:Hinton:2015}
\textsc{LeCun, Yann}; \textsc{Yoshua Bengio}; and \textsc{Geoffrey~E. Hinton}.
  2015.
\newblock Deep learning.
\newblock \emph{Nature} 521.436--444.

\bibitem[{Lenci(2018)}]{Lenci:2018}
\textsc{Lenci, Allesandro}. 2018.
\newblock Distributional models of word meaning.
\newblock \emph{Annual Review of Linguistics} 4.151--171.

\bibitem[{Lewis(1970)}]{Lewis70GS}
\textsc{Lewis, David}. 1970.
\newblock General semantics.
\newblock \emph{Synthese} 22.18--67.

\bibitem[{Liang \& Potts(2015)Liang and Potts}]{Liang:Potts:2015}
\textsc{Liang, Percy}, and \textsc{Christopher Potts}. 2015.
\newblock Bringing machine learning and compositional semantics together.
\newblock \emph{Annual Review of Linguistics} 1.355--376.

\bibitem[{Manning(2015)}]{Manning:2015}
\textsc{Manning, Christopher~D.} 2015.
\newblock Computational linguistics and deep learning.
\newblock \emph{Computational Linguistics} 41.701--707.

\bibitem[{Mitchell \& Lapata(2010)Mitchell and Lapata}]{Mitchell:Lapata:2010}
\textsc{Mitchell, Jeff}, and \textsc{Mirella Lapata}. 2010.
\newblock Composition in distributional models of semantics.
\newblock \emph{Cognitive Science} 34.1388--1429.

\bibitem[{Montague(1970)}]{Montague70UG}
\textsc{Montague, Richard}. 1970.
\newblock Universal grammar.
\newblock \emph{Theoria} 36.373--398, Reprinted in \citet{Montague74},
  222--246.

\bibitem[{Montague(1974)}]{Montague74}
\textsc{Montague, Richard}. 1974.
\newblock \emph{Formal philosophy: Selected papers of {R}ichard {M}ontague}.
\newblock New Haven, CT: Yale University Press.

\bibitem[{Partee(1980)}]{Partee:1980}
\textsc{Partee, Barbara~H.} 1980.
\newblock Semantics -- mathematics or psychology?
\newblock \emph{Semantics from different points of view}, ed.~by Egli
  B{\"a}uerle and Arnim von Stechow, 1--14. Berlin: Springer-Verlag.

\bibitem[{Partee(1981)}]{Partee81}
\textsc{Partee, Barbara~H.} 1981.
\newblock Montague grammar, mental representations, and reality.
\newblock \emph{Philosophy and grammar}, ed.~by Stig Kanger and Sven
  {\"{O}}hman, 59--78. Dordrecht: D.~Reidel.

\bibitem[{Partee(1984)}]{Partee84}
\textsc{Partee, Barbara~H.} 1984.
\newblock Compositionality.
\newblock \emph{Varieties of formal semantics}, ed.~by Fred Landman and Frank
  Veltman, 281--311. Dordrecht: Foris.

\bibitem[{Partee(1995)}]{Partee95}
\textsc{Partee, Barbara~H}. 1995.
\newblock Lexical semantics and compositionality.
\newblock \emph{Invitation to cognitive science}, ed.~by Lila~R. Gleitman and
  Mark Liberman, vol.~1, 311--360. Cambridge, MA: MIT Press.

\bibitem[{Pater(2018)}]{Pater:2018}
\textsc{Pater, Joe}. 2018.
\newblock Generative linguistics and neural networks at 60: Foundation,
  friction, and fusion.
\newblock To appear in \emph{Language}.

\bibitem[{Plate(1994)}]{Plate:1994}
\textsc{Plate, Tony~A.} 1994.
\newblock \emph{Distributed representations and nested compositional
  structure}.
\newblock Toronto: University of Toronto dissertation.

\bibitem[{Pollack(1990)}]{Pollack:1990}
\textsc{Pollack, Jordan~B.} 1990.
\newblock Recursive distributed representations.
\newblock \emph{Artificial Intelligence} 46.77--105.

\bibitem[{Potts \& Levy(2015)Potts and Levy}]{Potts:Levy:2015}
\textsc{Potts, Christopher}, and \textsc{Roger Levy}. 2015.
\newblock Negotiating lexical uncertainty and speaker expertise with
  disjunction.
\newblock \emph{Proceedings of the 41st annual meeting of the {B}erkeley
  {L}inguistics {S}ociety}, ed.~by Anna~E. Jurgensen, Hannah Sande, Spencer
  Lamoureux, Kenny Baclawski, and Alison Zerbe, 417--445. Berkeley, CA:
  Berkeley Linguistics Society.

\bibitem[{Ruppenhofer et~al.(2006)Ruppenhofer and colleagues}]{FrameNet2}
\textsc{Ruppenhofer, Josef}; \textsc{Michael Ellsworth}; \textsc{Miriam R.~L.
  Petruck}; \textsc{Christopher~R. Johnson}; and \textsc{Jan Scheffczyk}. 2006.
\newblock \emph{Frame{N}et {II}: Extended theory and practice}.
\newblock Berkeley, CA: International Computer Science Institute.

\bibitem[{Searle(1972)}]{Searle:1972}
\textsc{Searle, John~R.} 1972.
\newblock Chomsky's revolution in linguistics.
\newblock \emph{The New York Review of Books} 18.12--29.

\bibitem[{Smolensky(1990)}]{Smolensky:1990}
\textsc{Smolensky, Paul}. 1990.
\newblock Tensor product variable binding and the representation of symbolic
  structures in connectionist systems.
\newblock \emph{Artificial Intelligence} 46.159--216.

\bibitem[{Socher et~al.(2012)Socher and colleagues}]{Socher-etal:2012}
\textsc{Socher, Richard}; \textsc{Brody Huval}; \textsc{Christopher~D.
  Manning}; and \textsc{Andrew~Y. Ng}. 2012.
\newblock Semantic compositionality through recursive matrix-vector spaces.
\newblock \emph{Proceedings of the 2012 {J}oint {C}onference on {E}mpirical
  {M}ethods in {N}atural {L}anguage {P}rocessing and {C}omputational {N}atural
  {L}anguage {L}earning}, Jeju Island, Korea, 1201--1211.
\urlprefix\url{http://www.aclweb.org/anthology/D12-1110}.

\bibitem[{Socher et~al.(2011)Socher and colleagues}]{Socher-etal:2011:EMNLP}
\textsc{Socher, Richard}; \textsc{Jeffrey Pennington}; \textsc{Eric~H. Huang};
  \textsc{Andrew~Y. Ng}; and \textsc{Christopher~D. Manning}. 2011.
\newblock Semi-supervised recursive autoencoders for predicting sentiment
  distributions.
\newblock \emph{Proceedings of the 2011 conference on {E}mpirical {M}ethods in
  {N}atural {L}anguage {P}rocessing}, 151--161. Edinburgh, Scotland, UK: ACL.
\urlprefix\url{http://www.aclweb.org/anthology/D11-1014}.

\bibitem[{Socher et~al.(2013)Socher and colleagues}]{Socher-etal:2013}
\textsc{Socher, Richard}; \textsc{Alex Perelygin}; \textsc{Jean Wu};
  \textsc{Jason Chuang}; \textsc{Christopher~D. Manning}; \textsc{Andrew~Y.
  Ng}; and \textsc{Christopher Potts}. 2013.
\newblock Recursive deep models for semantic compositionality over a sentiment
  treebank.
\newblock \emph{Proceedings of the 2013 conference on {E}mpirical {M}ethods in
  {N}atural {L}anguage {P}rocessing}, 1631--1642. Seattle, WA: Association for
  Computational Linguistics.
\urlprefix\url{http://www.aclweb.org/anthology/D13-1170}.

\bibitem[{Thomason(1974)}]{Thomason74INTRO}
\textsc{Thomason, Richmond~H.} 1974.
\newblock Introduction.
\newblock In \citeauthor{Montague74}, 1--69.

\bibitem[{Turney \& Pantel(2010)Turney and Pantel}]{TurneyPantel10}
\textsc{Turney, Peter~D.}, and \textsc{Patrick Pantel}. 2010.
\newblock From frequency to meaning: Vector space models of semantics.
\newblock \emph{Journal of Artificial Intelligence Research} 37.141--188.

\bibitem[{Wilson \& Carston(2007)Wilson and Carston}]{Wilson:Carston:2007}
\textsc{Wilson, Dierdre}, and \textsc{Robyn Carston}. 2007.
\newblock A unitary approach to lexical pragmatics: Relevance, inference and ad
  hoc concepts.
\newblock \emph{Pragmatics}, ed.~by Noel Burton-Roberts, 230--259. Basingstoke
  and New York: Palgrave Macmillan.

\end{thebibliography}

\end{document}